\pdfoutput=1
\documentclass[a4paper]{article}

\usepackage{INTERSPEECH2021}
\usepackage[table,xcdraw]{xcolor}
\usepackage{soul}
\usepackage{url}
\usepackage{subfigure}
\usepackage{dblfloatfix} 

\title{Voice Quality and Pitch Features in Transformer-Based Speech Recognition}
\name{Guillermo C\'ambara$^{1,2}$, Jordi Luque$^{2}$ and Mireia Farr\'us$^3$}
\address{$^1$TALN Research Group, Universitat Pompeu Fabra, Barcelona, Spain\\
$^2$Telef\'onica I+D, Research, Barcelona, Spain\\
$^3$Language and Computation Centre, Universitat de Barcelona, Spain}
\email{guillermo.cambara@upf.edu, jordi.luque@telefonica.com, mfarrus@ub.edu}

\begin{document}

\maketitle
\begin{abstract}
Jitter and shimmer measurements have shown to be carriers of voice quality and prosodic information which enhance the performance of tasks like speaker recognition, diarization or automatic speech recognition (ASR). However, such features have been seldom used in the context of neural-based ASR, where spectral features often prevail. In this work, we study the effects of incorporating voice quality and pitch features altogether and separately to a Transformer-based ASR model, with the intuition that the attention mechanisms might exploit latent prosodic traits. For doing so, we propose separated convolutional front-ends for prosodic and spectral features, showing that this architectural choice yields better results than simple concatenation of such pitch and voice quality features to mel-spectrogram filterbanks. Furthermore, we find mean Word Error Rate relative reductions of up to 5.6\% with the LibriSpeech benchmark. Such findings motivate further research on the application of prosody knowledge for increasing the robustness of Transformer-based ASR.
\end{abstract}

\section{Introduction}

Research during the last years has shown success of neural network models applied to automatic speech recognition (ASR), consistently achieving state-of-the-art results in the field. The architectural choices vary within a wide range, from hybrid DNN-HMM models \cite{vesely2013sequence} to fully neural end-to-end approaches, like convolutional models  benefiting from gated linear units (GLUs) \cite{zeghidour2018fully, Liptchinsky2017LetterBasedSR, convGLU}, ResNet blocks \cite{wang2017residual, synnaeve2019endtoend} or time-depth separable (TDS) convolutions \cite{synnaeve2019endtoend, hannun2019sequence}.

The mentioned convolutional architectures are efficient approaches in the sense that they achieve low word error rate (WER) scores, while being very suitable for online streaming applications \cite{pratap2020scaling}. This effectiveness is due to their ability to extract relevant local features, although other recently proposed models seem to have an easier time capturing information from larger contexts. These are sequence-to-sequence models based on the attention mechanism \cite{bahdanau2014neural}, which is able to focus on the relevant parts of a whole sequence in order to generate the correct output, like the LAS model \cite{las2015}. Actually, many state-of-the-art proposals mainly consist of Transformer blocks \cite{vaswani2017attention}, which solely rely on the attention mechanism. Transformer models might be complemented with convolutional operations at their front-ends \cite{synnaeve2019endtoend, mohamed2019transformers}, or even within the Transformer blocks themselves, a structure known as Conformer \cite{gulati2020conformer}. This way, local feature extraction is enhanced, an important factor knowing that attention acts by relating time frames in the signal, each one of them represented by an acoustic feature vector, in a similar way as words are encoded in embeddings in the NLP field.

Currently, most of the modern architectures work only on cepstral and mel-frequency spectral coefficients inputs, or even directly with the raw waveform, tending towards the increasing of granularity at input level by usually augmenting the number of spectral parameters. Whilst it seems evident that current end-to-end deep architectures are able to automatically perform relevant feature extraction for speech tasks, psychical or functional properties, related to the underlying speech production system, become fuzzy or difficult to connect with the speech recognition performances. In addition, it is still unclear how the great quantity and different speech hand-crafted voice features, carefully developed along past years and based on our linguistic knowledge, might help and in which degree to the current ASR neural architectures. To name a few, pitch --or fundamental perceived frequency--, is an important prosodic information carrier, or jitter and shimmer, which represent the cycle-to-cycle variations of the fundamental frequency and amplitude, respectively, and are related to the quality of speech. 

The main contribution of this work is assessing the value of two ways of adding voice quality (VQ) and pitch features to the spectral coefficients, commonly employed in most neural ASR systems. Firstly, by simple concatenation before the network's forward pass. Secondly, by using two separated convolutional front-ends, one for spectral features and another for voice quality plus pitch features, concatenating the outcome features from the front-ends. Experiments are carried out with the LibriSpeech dataset \cite{librispeech}, using the Transformer-based model \cite{vaswani2017attention} from fairseq's toolkit speech-to-text (S2T) recipe \cite{ott2019fairseq, wang2020fairseq}. Up to our knowledge, this is the first attempt to use jitter and shimmer features within a modern deep neural-based speech recognition system 
while keeping {\it easy to identify} psychical/functional properties of the voice and linking them to the ASR performance. Furthermore, the recipe employed in this research is made available for the community in a GitHub\footnote{\url{https://github.com/gcambara/speechbook/tree/master/recipes/vq_pitch}} repository. We hypothesize that prosodic and voice quality features may boost robustness in state-of-the-art neural ASR systems like the Transformer-based model used in this paper. Acoustic spectral feature vectors would be enrichened with such information, in our case specifically using jitter and shimmer local variants, the best performing ones in \cite{farrus2007jitter}. These might complement pitch features or even be beneficial on themselves.

The structure of the paper is as follows. Section \ref{related_work} briefly overviews the related work on pitch and voice quality features, Sections \ref{model_description} and \ref{methodology} describe the model and methodology used in the current work, respectively, Section \ref{results} shows the results obtained, and Section \ref{conclusions} illustrates the final conclusions.

\section{Pitch and VQ Features in ASR}\label{related_work}

Some well-known speech recognition frameworks, like Kaldi \cite{kaldi}, have incorporated the use of additional prosodic features, such as the pitch or the probability of voicing (POV). These are stacked into an input vector together with those cepstral/spectral ones, and then forwarded to classifiers like HMM or DNN ensembles. Such features have proved to increase performance in ASR systems, significantly for tonal languages, like Punjabi \cite{GUGLANI2020107386}, as well as for non-tonal ones like English \cite{pitchASR}. Nevertheless, prosodic features are seldom used in the context of Transformer-based or convolutional speech recognition, often relying on the sole use of spectral features. 

Regarding jitter and shimmer measurements, they have been considered relevant and often applied to detect voice pathologies for a long time, as in \cite{kreiman2005perception,michaelis1998some}. Although voice quality features differ intrinsically from suprasegmental prosodic ones like pitch itself, they have shown to be related to prosody. In \cite{campbellvoice}, the authors showed that voice quality features are relevant markers signaling paralinguistic information, and that they should even be considered as prosodic carriers along with pitch and duration, for instance.
Furthermore, in the last decades, jitter and shimmer have shown to be useful in a wide range of applications; e.g. detection of different speaking styles \cite{slyh1999analysis}, age and gender classification \cite{wittig2003implicit}, emotion detection \cite{li2007stress}, speaker recognition \cite{farrus2007jitter}, speaker diarization \cite{zewoudie2014jitter}, and Alzheimer's and Parkinson Diseases detection \cite{mirzaei2018two,benba2014hybridization}, among others. 

In ASR literature, some works have reported that prosodic information can raise the performance of speech recognizer systems. For instance, in \cite{zaidi2019automatic} the authors built an ASR for dysarthric speech and \cite{robustJitter} reports benefit on the use of jitter and shimmer for noisy speech recognition, both systems based on classical HMM acoustic modelling. Actually, our choice of processing prosodic and spectral features separately before fusion has been inspired by other works in speaker verification \cite{soleymani2018prosodic} and speaker recognition \cite{farrus2006fusion}. In our previous work \cite{cambara2020convolutional}, we employed voice quality features in a convolutional-based ASR, showing only slight improvements in WER. Besides such work, and up to our knowledge, voice quality features have not been used yet for neural-based ASR. Since Transformer models use attention to find correlations in larger contexts, we think that prosodic information from pitch and voice quality features could be better captured compared to convolutional approaches, due to its suprasegmental nature.

\begin{figure}[h]
  \centering
  \includegraphics[width=\linewidth]{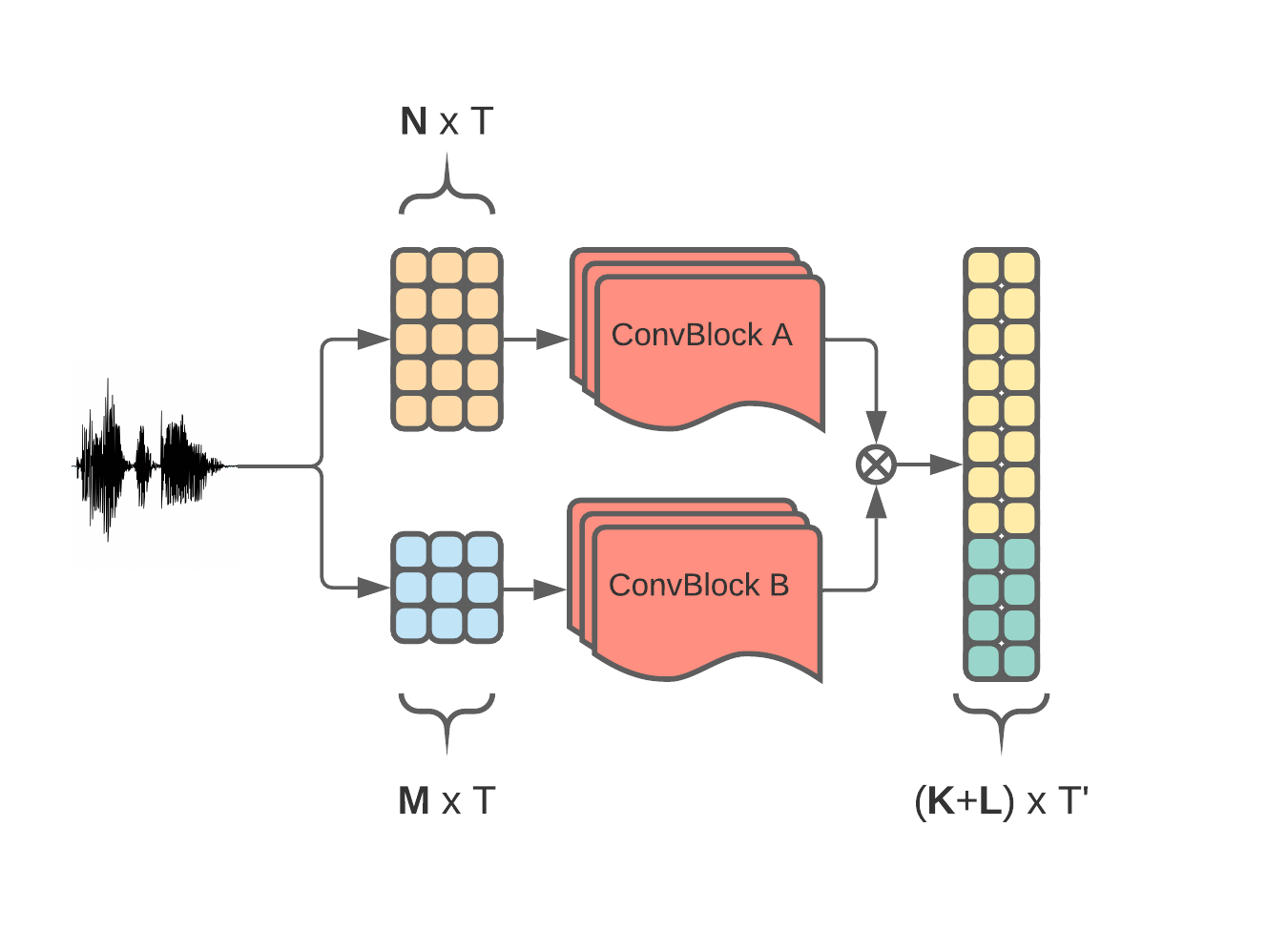}
  \caption{Convolutional front-end for mel-spectrogram filterbanks and pitch related features in S2T VQ Transformer  architecture.}
  \label{fig:vq_arch}
  \vspace{-1.5em}
\end{figure}

\section{Model Description}\label{model_description}
The addition of voice quality and pitch features to spectral coefficients is tackled early on, in the front-end of a Transformer-based model. The main structure of such model, named as S2T Transformer from now on, consists of a convolutional block as a front-end, followed by sinusoidal positional embedding of features and an encoder-decoder constituted by Transformer blocks, in the likes of \cite{vaswani2017attention}. The convolutional block is formed by two 1-dim convolutional layers with kernel size 5, stride 2 and padding 2, each followed by GLU activation functions \cite{convGLU}. The first layer takes $T$ time frames per $N=40$ mel-spectrograms as input, upsampling them to $1024$ features that are halved by the GLU activation, resulting in $P=512$ hidden features. These are passed to the second layer, which outputs the same feature dimension of $512$, yielding $O=256$ features after GLU activation, which is the embedding size of all the following attention layers.

Incorporating any combination of voice quality and pitch features with size $M \in [1, 5]$ to the mel-spectrograms causes the input feature vector to be of size $N+M$. Thus, the implicit pitch-related information is finally contained within the output feature vector of size $O=256$. However, we wonder whether such information could be too diluted after convolving with the mel-spectrograms, being $N>>M$. 

\begin{figure*}[b]
\centering     
\hspace{-0.3cm}
\begin{subfigure}{\label{fig:libri_100h_testclean}\includegraphics[width=0.45\linewidth]{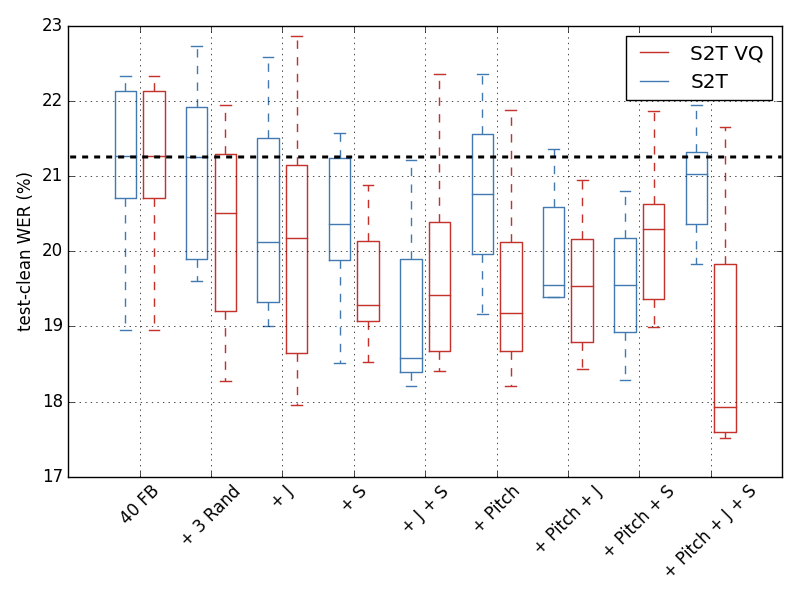}}
\hspace{-0.35cm}
\end{subfigure}
\subfigure{\label{fig:libri_100h_testother}\includegraphics[width=0.45\linewidth]{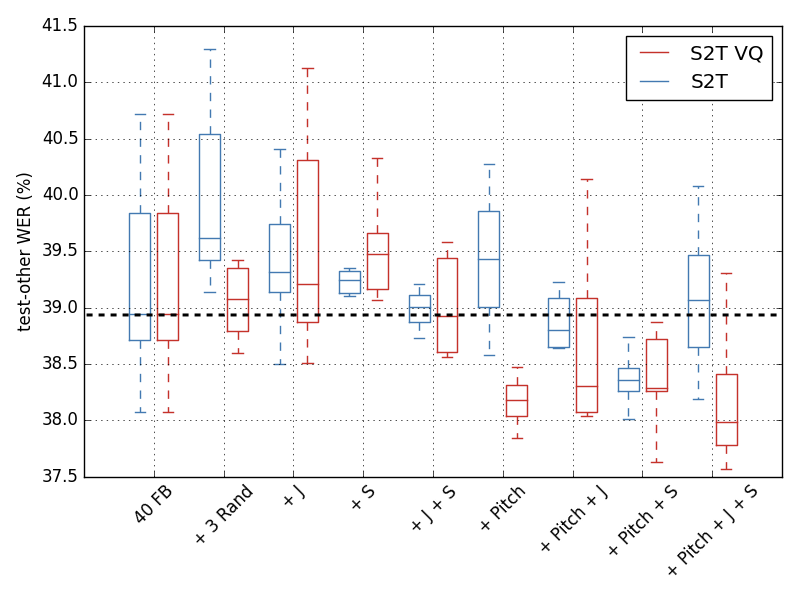}}
\caption{LibriSpeech test-clean and test-other WER (\%) for several feature configurations tested, combining mel-spectrogram filterbanks (FB), 3 random features (Rand), F0+POV+$\Delta$F0 (Pitch), jitter (J) and shimmer (S), using S2T and S2T VQ models. 40 FB baseline is marked by the dashline.}
\vspace{-0.6cm}
\end{figure*}


For such reason, we propose an architectural variant called S2T VQ Transformer, shown in Figure \ref{fig:vq_arch}, which consists of two convolutional blocks, $A$ and $B$, receiving mel-spectrograms and voice quality plus pitch features, respectively. These are convolved independently and concatenated afterwards, so the resulting feature vector contains a representation of $K$ spectral features and $L$ pitch-related features, yielding a total output size of $O' = K + L$. By controlling the proportion between $K$ and $L$ we decide how much weight we assign to the new features, so the attention layers may take these more into account. Each block $A$ and $B$ is identical to the block in the plain S2T Transformer architecture, except for the variations showed in Table \ref{tab:conv_frontends}.


\begin{table}[h]
\caption{Convolutional front-end blocks for mel-spectrograms and voice quality + pitch features.}
\label{tab:conv_frontends}
\centering
\begin{tabular}{lccc}
\toprule
\textbf{Block} & \textbf{Input Dim} &  \textbf{Hidden Dim} & \textbf{Output Dim} \\
\midrule
A & $N=40$  & $p_A=512$ & $K=192$ \\
B & $M \in [1, 5]$  & $p_B=256$ & $L=64$ \\
\bottomrule
\end{tabular}
\end{table}


Note that the hidden dimension $p_A$ is kept for the spectral block $A$, but for block $B$ is reduced to the half, since the amount of pitch and voice quality features is smaller and probably we do not need as much up-sampling. By setting output dimensions for each front-end at $K=192$ and $L=64$, concatenation of these yields the same total output size as plain S2T Transformer model: $O'=O=256$. This way, we ensure fairness in the comparison between plain S2T and VQ S2T models, since the simple fact of increasing $O'$ might yield better results due to the increased number of parameters in the feature vectors, maybe better exploited by attention. Actually, the VQ variant has slightly less parameters, 29.2M against 29.4M in the plain model. Furthermore, this choice keeps a proportion of 1 pitch-related feature for every 3 spectral features, giving a larger weight to the first ones compared to the plain S2T Transformer model.

\section{Methodology}
\label{methodology}

\subsection{Dataset}

The proposal of this work is assessed with the LibriSpeech dataset \cite{librispeech}, which contains up to 1000 hours of audio book recordings in English, sampled at 16kHz. 
LibriSpeech was deployed with a ready-to-go train split subdivided into three subsets with approximately 100, 360 and 500 hours of speech. For development and test sets, two subsets were created as well. Hereinafter, the aforementioned partitions shall be referred as train-clean-100, train-clean-360, train-other-500, dev-clean, dev-other, test-clean and test-other, respectively.

\subsection{Experimental setup}

Aiming to explore the information conveyed by pitch and voice quality features in the speech recognition system, we mainly train and test independent acoustic models using different feature combinations: mel-spectrograms, fundamental frequency (F0), POV, $\Delta$F0, jitter and shimmer. To ensure that additional voice quality and pitch features do not cause a positive impact simply because they increase the number of parameters in the model, we also try a configuration of 3 features generated with random numbers sampled from a uniform distribution between 0 and 10, similar to mel-spectrogram filterbanks. Performance for every model is evaluated in terms of WER in test-clean and test-other sets. The whole train and test procedure is repeated for every feature configuration across 6 different initialization seeds, calculating the average WER plus the standard deviation of the mean to ensure statistical significance of the results. We decide to train with the train-clean-100 set only for the seed extensive experiments, due to computational constraints. Afterwards, we compare the performance of the filterbank-only baseline against the fully-packed S2T VQ model with pitch and voice quality features. For this, we explore different training set hours: 50, 100, 200, 500 and 960.   

\subsection{Acoustic modeling}

Training and testing of acoustic models is done with the fairseq toolkit \cite{ott2019fairseq}, which contains functionality and examples for the S2T task \cite{wang2020fairseq}. Specifically, we use their LibriSpeech ASR example as a starting point for our experiments, training the S2T small Transformer model (31M parameters) with mel-spectrogram features computed on-the-fly by fairseq. 

For experiments requiring pitch and voice quality features, we precompute these with Praat-Parselmouth \cite{jadoul2018introducing}, a Python wrapper for Praat \cite{boersma2001speak}, a well tested library for speech processing and analysis. We compute all features with a window size of 25 ms and a stride of 10 ms. The pitch, jitter and shimmer extraction algorithms yield empty values for unvoiced frames, so following Kaldi's pitch extraction algorithm recommendation \cite{Ghahremani2014APE}, we interpolate these with the adjacent non-empty values. We keep the empty indexes to construct a POV vector where voiced and unvoiced segments are denoted by $1$ and $-1$, respectively. The logarithm is applied to the pitch vector and pitch, jitter and shimmer vectors are smoothed out by computing the mean with a centered window across 151 frames, in the style of Kaldi. Finally, all the combined features are normalized by cepstral mean and variance normalization.

From this point, features may be handled in two different ways, depending on the chosen front-end, as described in section \ref{model_description}: simply stacked with mel-spectrograms, or processed in a dedicated convolutional block and then stacked with the convolved mel-spectrograms. We use 40 filterbanks, instead of the 80 filterbanks in the original fairseq example, in order to increase the proportion of pitch and voice quality features in the total amount of features. This is done to yield a higher numerical importance to the new set of features, specially for the simple stacking experiments, while keeping a commonly used number of filterbanks. Since simple stacking configuration is not used for the training set size experiments, we then switch back to 80 filterbanks.

Models are trained using the cross entropy criterion with label smoothing and Adam optimization, clipping gradient values above 10.0 to prevent gradient explosions. The learning rate peaks at 0.002 after warming up for 10k batch updates, being decreased afterwards by an inverse root scheduler. For the experiments regarding the different training sizes, we use the exact hyperparameters of the fairseq recipe, to be closer to its baseline. We leave them training for many iterations after validation loss plateaus, to ensure convergence: 300k updates for the 960 hours model, 150k updates for the 500, 200 and 100 hours ones, and 55k for the 50 hours model. However, we restrict the number of iterations to 20k batch updates for the feature configuration scan across seeds, since this is computationally costly and no significant gains are observed for longer trainings.




\begin{table}[b!]
\renewcommand{\arraystretch}{1.50}
\caption{Mean test WER scores for acoustic models trained with different subset sizes of LibriSpeech training. Two architectures are used: the baseline with filterbank features (80 FB) and the S2T VQ model with filterbank, pitch and voice quality features (+VQ).}
\label{libri_hours_table}
\small{
\begin{tabular*}{\columnwidth}{l @{\extracolsep{\fill}} cccc}
\hline
  \multicolumn{1}{c}{Train set}  & \multicolumn{4}{c}{WER (\%)}   \\
\hline
\bfseries Hours & \multicolumn{2}{c}{\bfseries test-clean} & \multicolumn{2}{c}{\bfseries test-other} \\
 & \bfseries 80 FB & \bfseries + VQ & \bfseries 80 FB & \bfseries + VQ\\
\hline

\hline\hline
50 & $29.26 $ & $\mathbf{28.30}$ & $47.70$ & $\mathbf{46.91}$ \\
100 & $19.33$ & $\mathbf{18.29}$ & $37.04$ & $\mathbf{35.31}$\\
200 & $10.95$ & $\mathbf{10.80}$ & $\mathbf{24.94}$ & $25.32$\\
500 & $7.78$ & $\mathbf{6.96}$ & $18.34$ & $\mathbf{17.56}$\\
960 & $4.62$ & $\mathbf{4.27}$ & $10.64$ & $\mathbf{10.01}$\\

\hline
\end{tabular*}
\vspace{0.1cm}

}
\end{table}

\subsection{Decoding}

We found better WER scores by averaging the weights from the last 10 checkpoints in training, which is the method in the fairseq recipe, rather than selecting the checkpoint with best development WER. Decoding is done by means of the beam search algorithm with beam size 5, using a 10k unigram lexicon extracted with sentencepiece \cite{kudo2018sentencepiece} from the LibriSpeech corpus. Since the aim of this work is the assessment of voice quality and pitch features for acoustic modeling, no additional language model is used.

\vspace{-1em}
\section{Results}
\label{results}

Figures \ref{fig:libri_100h_testclean} and \ref{fig:libri_100h_testother} depict the WER score distributions for a selection of the feature combinations across the 6 different seeds employed, for the LibriSpeech test sets. Overall, mean WER scores for the S2T VQ model tend to be smaller than plain S2T Transformer using simple concatenation of features. Furthermore, there is a trend in WER reduction as the number of added features increases, specially for the S2T VQ architecture. This is particularly notable when compared to the baseline of 40 filterbanks, or the model using 3 random features.

Pitch and VQ features seem to benefit from an architecture using specific convolutions for them, as hypothesized. Jitter and shimmer only slightly improve the performance of the system, achieving reductions in mean WER of $1.3\%$ and $0.2\%$ for test-clean and test-other, respectively. Pitch achieves a better performance, improving $1.5\%$ and $1.0\%$ mean WER for the same sets. However, the biggest improvement comes from joining both pitch plus jitter and shimmer features. Mean WER reduction at test-clean is around $2.3\%$, whereas for test-other is $1.0\%$. Two tailed p-values in the pitch experiments are $0.134$ and $0.069$, showing a fair significance, whereas the jitter and shimmer experiments yield low significance, with p-values of $0.129$ and $0.624$. Nevertheless, the experiments with pitch and VQ features altogether show a relevant significance, with p-values of $0.027$ and $0.052$ for test-clean and test-other. This suggests that jitter and shimmer might be a good complement to improve pitch features, although they alone do not cause a significant change.

\begin{figure}[t]
  \centering
  \includegraphics[width=\linewidth]{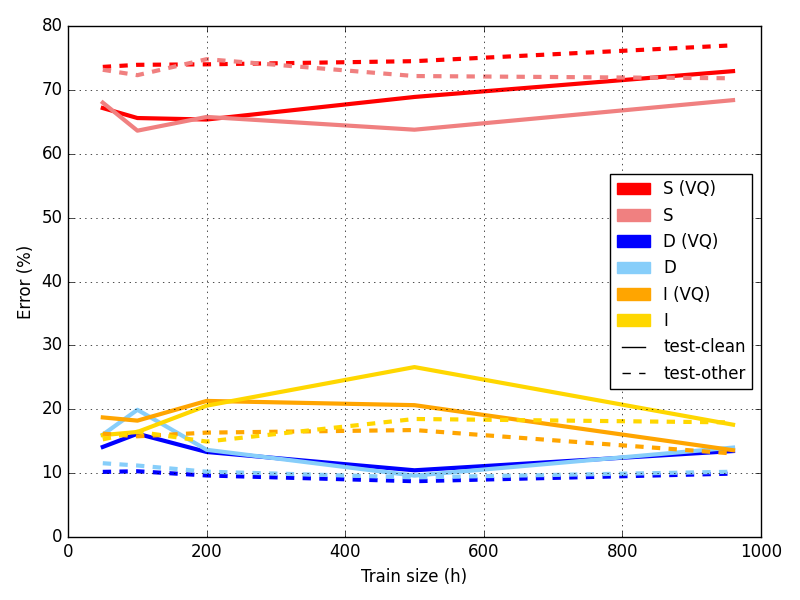}
  \caption{Error distributions across training hours. Error types: substitutions (S), deletions (D) and insertions (I).}
  \label{fig:mean_error_type}
  \vspace{-1.5em}
\end{figure}

Table \ref{libri_hours_table} shows that the fully-packed S2T VQ model surpasses the filterbank-only baseline in the majority of training size setups. The relative WER reduction averaged across training setups is $5.6\%$ for test-clean and $3.0\%$ for test-other, suggesting that the benefits of using pitch and VQ features hold as the number of train hours increases. The influence of using pitch and VQ features as training hours grow is better understood looking at the error type distributions in Figure \ref{fig:mean_error_type}. The evolution of deletions (D) is very similar in both cases, but the S2T VQ model is significantly better at reducing the amount of insertions (I), leaving a higher representation of substitutions (S). This suggests that, with enough data, the S2T VQ model might learn to better leverage prosody information for delimiting beginning and ending of words.

\section{Conclusions}
\label{conclusions}

The incorporation of pitch and voice quality features to  spectral features has been studied for a Transformer-based state-of-the-art acoustic ASR, in two ways: the first being simple concatenation, and the second one exploring concatenation after separately convolving such features in two separated front-ends. The latter proposal clearly outperforms the former, showing WER improvements by using pitch and voice quality features separately, significantly yielding the best performances when used together, with up to mean $5.6\%$ relative WER reductions, specially taming insertion errors. Thus, results show the potential of voice quality features for complementing prosodic information carried in pitch features. Using separated convolutional filters for these features increases the representation of such prosodic information in the acoustic features, which suprasegmental nature might be better leveraged by the attention layers. Besides further explorations with bigger and more diverse data sets, we think that the findings of this work motivate further research in conversational speech recognition or speech recognition with punctuation signs, where prosody plays an important role. 

\section{Acknowledgements}

This work is a part of the INGENIOUS project, funded by the European Union’s Horizon 2020 Research and Innovation Programme and the Korean Government under Grant Agreement No 833435. The third author has been funded by the Agencia Estatal de Investigación (AEI), Ministerio de Ciencia, Innovación y Universidades and the Fondo Social Europeo (FSE) under grant RYC-2015-17239 (AEI/FSE, UE).

\bibliographystyle{IEEEtran}

\bibliography{mybib}


\end{document}